\documentclass[lettersize,journal]{IEEEtran}
\usepackage{amsmath,amsfonts}
\usepackage{array}
\usepackage[caption=false,font=normalsize,labelfont=sf,textfont=sf]{subfig}
\usepackage{textcomp}
\usepackage{stfloats}
\usepackage{url}
\usepackage{verbatim}
\usepackage{graphicx}
\usepackage{cite}
\usepackage{multirow}
\usepackage{enumitem}
\usepackage{makecell}
\usepackage[noend]{algpseudocode}
\usepackage{algorithm}
\usepackage{soul}
\usepackage{xcolor}
\usepackage{hyperref}
\begin{document}

\title{Dual Accuracy-Quality-Driven Neural Network for Prediction Interval Generation}

\author{Giorgio Morales,~\IEEEmembership{Member,~IEEE,} and John W. Sheppard,~\IEEEmembership{Fellow,~IEEE}
\thanks{Manuscript received November 1, 2022; revised March  13, 2023; revised August 3, 2023; accepted December 1, 2023. 
This research was supported by a USDA-NIFA-AFRI Food Security Program Coordinated Agricultural Project 
(Accession Number 2016-68004-24769), and also by the USDA-NRCS Conservation Innovation Grant from the On-farm Trials Program
(Award Number NR213A7500013G021).}
\thanks{Giorgio Morales and John W. Sheppard are with the Gianforte School of Computing, Montana State University, Bozeman, MT 59717, US (email: giorgiol.moralesluna@student.montana.edu; john.sheppard@montana.edu).}
\thanks{This paper is a preprint (submitted to IEEE Transactions on Neural Networks and Learning Systems). \copyright~\copyright~2022 IEEE. Personal use of this material is permitted. Permission from IEEE must be obtained for all other uses, in any current or future media, including reprinting/republishing this material for advertising or promotional purposes, creating new collective works, for resale or redistribution to servers or lists, or reuse of any copyrighted component of this work in other works}
}


\markboth{IEEE TRANSACTIONS ON NEURAL NETWORKS AND LEARNING SYSTEMS,~Vol.~XX, No.~X, December~2023}%
{Morales and Sheppard: Dual Accuracy-Quality-Driven Neural Network for Prediction Interval Generation}


\maketitle

\begin{abstract}
Accurate uncertainty quantification is necessary to enhance the reliability of deep learning models in real-world applications.
In the case of regression tasks, prediction intervals (PIs) should be provided along with the deterministic predictions of deep learning models.
Such PIs are useful or ``high-quality'' as long as they are sufficiently narrow and capture most of the probability density. 
In this paper, we present a method to learn prediction intervals for regression-based neural networks automatically in addition to the conventional target predictions.
In particular, we train two companion neural networks: one that uses one output, the target estimate, and another that uses two outputs, the upper and lower bounds of the corresponding PI.
Our main contribution is the design of a novel loss function for the PI-generation network that takes into account the output of the target-estimation network and has two optimization objectives: minimizing the mean prediction interval width and ensuring the PI integrity using constraints that maximize the prediction interval probability coverage implicitly.
Furthermore, we introduce a self-adaptive coefficient that balances both objectives within the loss function, which alleviates the task of fine-tuning.
Experiments using a synthetic dataset, eight benchmark datasets, and a real-world crop yield prediction dataset showed that our method was able to maintain a nominal probability coverage and produce significantly narrower PIs without detriment to its target estimation accuracy when compared to those PIs generated by three state-of-the-art neural-network-based methods.
In other words, our method was shown to produce higher-quality PIs.

\end{abstract}

\begin{IEEEkeywords}
Prediction intervals, companion networks, uncertainty quantification, deep regression.
\end{IEEEkeywords}

\section{Introduction} \label{sec:intro}

Deep learning has gained a great deal of attention due to its ability to outperform alternative machine learning methods in solving complex problems in a variety of domains. 
In conjunction with the availability of large-scale datasets and modern parallel hardware architectures (e.g., GPUs), convolutional neural networks (CNNs), as one popular deep learning technique, have attained unprecedented achievements in fields such as computer vision, speech recognition, natural language processing, medical diagnosis, and others \cite{reviewDL}.

While the undeniable success of deep learning (DL) has impacted applications that are used on a daily basis, 
many theoretical aspects remain unclear, which is why these models are usually referred to as ``black boxes'' in the literature~\cite{XAI}.
In addition, numerous reports suggest that current DL techniques typically lead to unstable predictions that can occur randomly and not only in worst-case scenarios~\cite{instability}.
As a consequence, they are considered unreliable for applications that deal with uncertainty in the data or in the underlying system, such as weather forecasting~\cite{wforecast}, electronic manufacturing~\cite{manufact}, 
or precision agriculture~\cite{PAreliable}. 
Note that, in this context, reliability is defined as the ability for a model to work consistently across real-world settings~\cite{tran2022plex}.

One of the limitations of conventional neural networks is that they only provide deterministic point estimates without any additional indication of their approximate accuracy~\cite{pmlr-v48-gal16}.
Reliability and accuracy of the generated point predictions are affected by factors such as the sparsity of training data or target variables affected by probabilistic events~\cite{PI-NNreview}.
One way to improve the reliability and credibility of such complex models is to quantify the uncertainty in the predictions they generate~\cite{SHRESTHA2006225}. 
This uncertainty ($\sigma^2_y$) can be quantified using prediction intervals (PIs), which provide an estimate of the upper and the lower bounds within which a prediction will fall according to a certain probability~\cite{LUBE}.
Hence, the amount of uncertainty for each prediction is provided by the width of its corresponding PI.
PIs account for two types of uncertainty: model uncertainty ($\sigma^2_{model}$) and data noise variance ($\sigma^2_{noise}$)~\cite{LUBE}, where $\sigma^2_y = \sigma^2_{model} + \sigma^2_{noise}$. Model uncertainty arises due to model selection, training data variance, and parameter uncertainty~\cite{ICML-2018-PearceBZN}.
Data noise variance measures the variance of the error between observable target values and the outputs produced by the learned models.

Recently, some NN-based methods have been proposed to solve the PI generation problem~\cite{LUBE,LUBEGA,LUBEPSO,ICML-2018-PearceBZN,piven,pmlr-v124-saleh-salem20a}.
These methods  aim to train NNs using loss functions that aim to balance at least two of the following three objectives: minimizing mean PI width, maximizing PI coverage probability, and minimizing the mean error of the target predictions. 
Although the aforementioned works have achieved promising results, there exist some limitations that need to be addressed.
For instance, they rely on the use of deep ensembles; however, training several models may become impractical when applied to complex models and large datasets~\cite{ensemblereview}. 
Furthermore, their performance is sensitive to the selection of multiple tunable hyperparameters whose values may differ substantially depending on the application.
Therefore, fine-tuning an ensemble of deep NNs becomes a computationally expensive task.
Finally, methods that generate PI bounds and target estimations simultaneously have to deal with a trade-off between the quality of generated PIs and the accuracy of the target estimations.

\begin{figure}
    \centering
    \includegraphics[width=8cm]{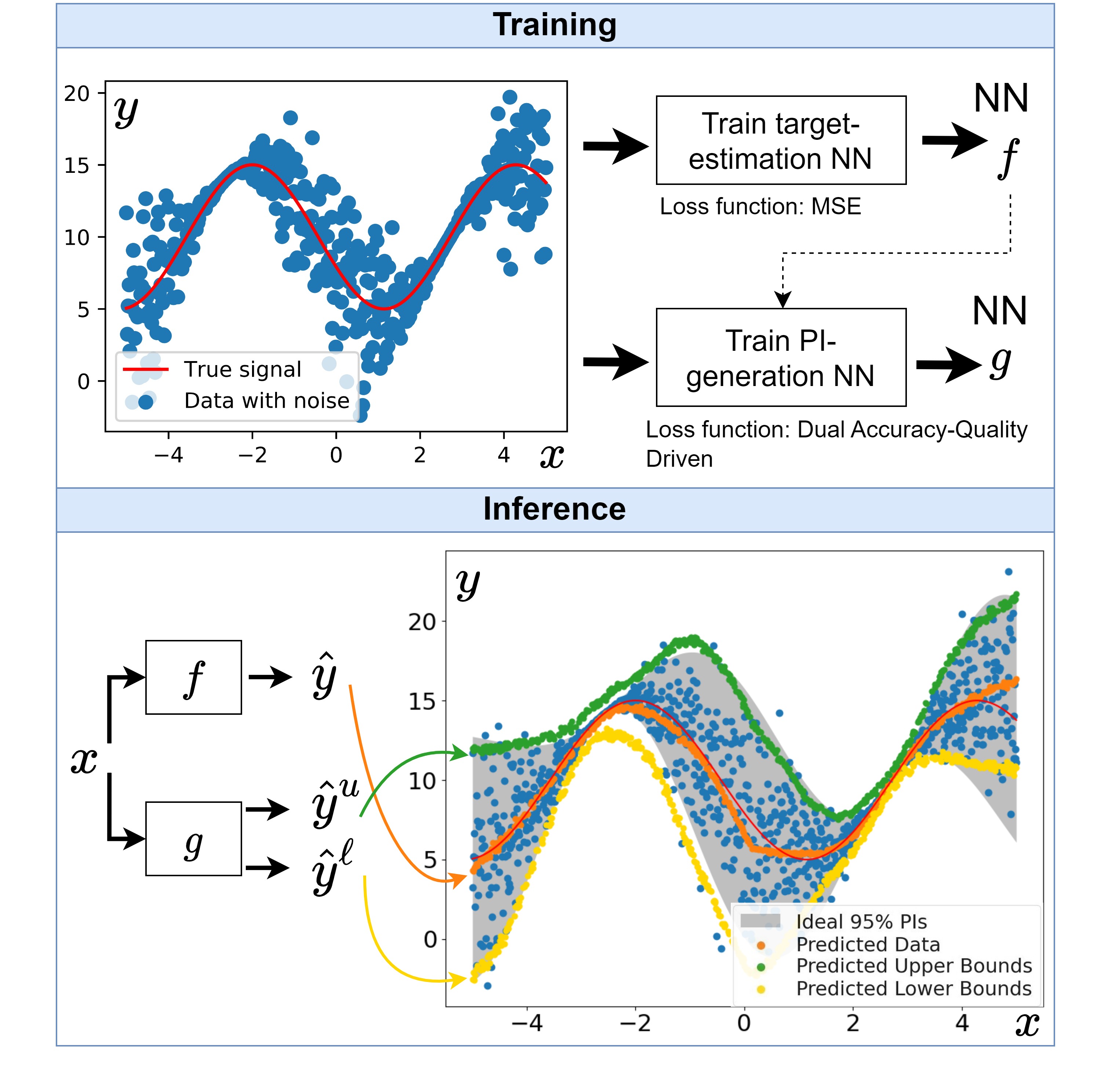}
    \vspace{-2ex}
    \caption{An example of our PI-generation method on a synthetic dataset.}
    \label{fig:intro}
\end{figure}

Pearce et al.~\cite{ICML-2018-PearceBZN} coined the term $\textit{High-quality (HQ)}$ $\textit{principle}$, which refers to the requirement that PIs be as narrow as possible while capturing some specified proportion of the predicted data points.   
Following this principle, we pose the PI generation problem for regression as a multi-objective optimization problem.
In particular, our proposal involves training two neural networks (NNs): one that generates accurate target estimations and one that generates narrow PIs (see Fig.~\ref{fig:intro}).

The first NN is trained to minimize the mean squared error of the target estimations.
Our main contribution is the design of a loss function for the second NN that, besides the generated PI bounds and the target, considers the output of the first NN as an additional input. 
It minimizes the mean prediction interval width and uses constraints to ensure the integrity of the generated PIs while implicitly maximizing the probability coverage (Sec.~\ref{sec:loss}).
Our second contribution is a method that updates the coefficient that balances the two optimization objectives of our loss function automatically throughout training (Sec.~\ref{sec:lambda}). 
Our method avoids generating unnecessarily wide PIs by using a technique that sorts the mini-batches at the beginning of each training epoch according to the width of the generated PIs (Sec.~\ref{sec:bs}).
Then we apply a Monte Carlo-based approach to account for the uncertainty of the generated upper and lower bounds
(Sec.~\ref{sec:MC}).
Finally, when compared to three state-of-the-art NN-based methods, we show that our method is able to produce PIs that maintain the target probability coverage while yielding better mean width without detriment to its target estimation accuracy (Sec.~\ref{sec:results}).

Our specific contributions are summarized as follows:
\begin{enumerate}
    \item Our main contribution is a novel loss function called Dual Accuracy-Quality-Driven (DualAQD) used to train a PI-generation NN. It is designed to solve a multi-objective optimization problem: minimizing the mean PI width while ensuring PI integrity using constraints that maximize the probability coverage implicitly.
    \item We present a new PI-generation framework that consists of two companion NNs: one that is trained to produce accurate target estimations, and another that generates high-quality PIs; thus, avoiding the common trade-off between target estimation accuracy, and quality of PIs.    
    \item We introduce a self-adaptive coefficient that balances the two objectives of our DualAQD loss function. This differs from previous approaches that consider this balancing coefficient as a tunable hyperparameter with a fixed value throughout the training process.  
    \item We present a method called batch-sorting that sorts the mini-batches according to their corresponding PI width and, as such, avoids generating unnecessarily wide PIs.
    \item Our method is shown to generate higher quality PIs and better reflects varying levels of uncertainty within the data than the compared methods.
\end{enumerate}


\section{Related Work} \label{background}

One of the more common approaches to uncertainty quantification for regression tasks is via Bayesian approaches, such as those represented by Bayesian neural networks (BNNs), 
which model the NN parameters as distributions.
As such, they have the advantage that they allow for a natural quantification of uncertainty.
In particular, uncertainty is quantified by learning a posterior weight distribution~\cite{neal2012bayesian,bnn}. 
The inference process involves marginalization over the weights, which in general is intractable, and sampling processes such as Markov chain Monte Carlo (MCMC) can be computationally prohibitive.
Thus, approximate solutions have been formulated using variational inference (VI)~\cite{VI}.
However, Wu et al.~\cite{Wu19} argued that VI approaches are fragile since they require careful initialization and tuning.
To overcome these issues, they proposed approximating moments in NNs to eliminate gradient variance.
They also presented an empirical Bayes procedure for selecting prior variances automatically.
Moreover, Izmailov et  al.~\cite{pmlr-v115-izmailov20a} discussed scaling BNNs to deep neural networks by constructing low-dimensional subspaces of the parameter space.
By doing so, they were able to apply elliptical slice sampling and VI, which struggle in the full parameter space.
In addition, Lut et al.~\cite{BSSCN} 
presented a Bayesian-learning-based sparse stochastic configuration network that replaces the Gaussian distribution with a Laplace one as the prior distribution for output weights.

Despite the aforementioned improvements in Bayesian approaches, they still suffer from various limitations.
Namely, the high dimensionality of the parameter space of deep NNs, including complex models such as CNNs, makes the cost of characterizing uncertainty over the parameters prohibitive~\cite{quality}.
Attempts to scale BNNs to deep NNs are considerably more expensive computationally than VI-based methods and have been scaled up to low-complexity problems only, such as MNIST~\cite{radialbnn}.
Conversely, non-Bayesian methods do not require the use of initial prior distributions and biases to train the models~\cite{LUBE}.
Recent works have demonstrated that non-Bayesian approaches provide better or competitive uncertainty estimates than their Bayesian counterparts~\cite{MVEEns,LUBE,ICML-2018-PearceBZN}.
In addition, they are scalable to complex problems and can handle millions of parameters.

MC-Dropout was proposed by Gal and Ghahramani~\cite{pmlr-v48-gal16} to quantify model uncertainty in NNs.
They cast dropout training in deep NNs as approximate Bayesian inference in deep Gaussian processes.
The method uses dropout repeatedly to select subsamples of active nodes in the network, turning a single
network into an ensemble.
Hence, model uncertainty is estimated by the sample variance of the ensemble predictions.
MC-Dropout is not able to estimate PIs themselves, as it does not account for data noise variance.
Therefore, Zhu and Laptev~\cite{MCPI} proposed estimating PIs by quantifying the model uncertainty through MC-Dropout, coupled with estimating the data noise variance as the mean squared error (MSE) calculated over an independent held-out validation set. 

Recently, several non-Bayesian approaches have been proposed for approximate uncertainty quantification.
Such approaches use models whose outputs provide estimations of the predictive uncertainty directly.
For instance, Schupbach et al.~\cite{ustatistics} proposed a method that estimates confidence intervals in NN ensembles based on the use of U-statistics.
Other techniques estimate PIs by using ensembles of feedforward networks~\cite{abbas} 
or stochastic configuration networks~\cite{SCN}
 and bootstrapping.
Lakshminarayanan et al.~\cite{MVEEns} presented an ensemble approach based on the Mean-Variance Estimation (MVE) method introduced by Nix and Weigend~\cite{MVE}. 
Here, each NN has two outputs: one that represents the mean (or target estimation) and the other that represents the variance of a normal distribution, which is used to quantify the data noise variance.
Other approaches use models that generate PI bounds explicitly.
Khrosavi et al.~\cite{LUBE} proposed a Lower Upper Bound Estimation (LUBE) method that uses a NN and a loss function to minimize the PI width while maximizing the probability coverage using simulated annealing.

Similar approaches have attempted to optimize the LUBE loss function using methods such as genetic algorithms~\cite{LUBEGA} and particle swarm optimization~\cite{LUBEPSO}.
Pearce et al.~\cite{ICML-2018-PearceBZN} proposed a method called QD-Ens that consists of a quality-driven loss function similar to LUBE but that is compatible with gradient descent.
Then Salem et al.~\cite{pmlr-v124-saleh-salem20a} proposed QD+ which is based on QD-Ens, which uses exactly the same two penalty functions to reduce the PI width and maximize the probability coverage. 
They used three-output NNs and included a third penalty term that aims to decrease the mean squared error of the target predictions and a fourth penalty term to enforce the point predictions to lay inside the generated PIs.
In our work, we use only three penalty terms; the differences are explained in Sec.~\ref{sec:comparison}.
Finally, both QD-Ens and QD+ used an ensemble approach to estimate the model uncertainty while we use a Monte Carlo approach on a single network.


\section{Proposed Methodology} \label{design}

\subsection{Dual Accuracy-Quality-Driven Loss Function} \label{sec:loss}

Let $\textbf{X}^b= \{ \textbf{x}_1, \dots , \textbf{x}_N \}$ be a training batch with $N$ samples where each sample $\textbf{x}_i \in \mathbb{R}^z$ consists of $z$ covariates.
Furthermore, let $\textbf{y}^b= \{ y_1, \dots , y_N \}$ be a  set of corresponding target observations where $y_i \in \mathbb{R}$.
We construct a NN regression model that captures the association between $\textbf{X}^b$ and $\textbf{y}^b$. 
More specifically, $f(\cdot)$ denotes the function computed by the NN, and $\boldsymbol{\theta}_f$ denotes its weights. 
Hence, given an input $\textbf{x}_i$, $f(\textbf{x}_i, \boldsymbol{\theta}_f)$ computes the target estimate $\hat{y}_i$.
This network is trained to generate accurate estimates $\hat{y}_i$ with respect to $y_i$. 
We quantify this accuracy by calculating the mean squared error of the estimation $
    MSE_{est} = \frac{1}{N} \sum_{i=1}^N (\hat{y}_{i} - y_{i})^2.$
Thus, $f$ is conventionally optimized as follows: 
\[
\boldsymbol{\theta}_f = \; \underset{\boldsymbol{\theta_f}}{\text{argmin}} \ MSE_{est}.
\label{eq:NNf}
\]

Once network $f(\cdot)$ is trained, we use a separate NN whose goal is to generate prediction intervals for $\textbf{y}^b$ given data $\textbf{X}^b$.
Let $g(\cdot)$ denote the function computed by this PI-generation NN, and $\boldsymbol{\theta}_g$ denotes its weights.
Given an input $\textbf{x}_i$, $g(\textbf{x}_i, \boldsymbol{\theta}_g)$ generates its corresponding upper and lower bounds, $\hat{y}^u_i$ and $\hat{y}^\ell_i$, such that $[\hat{y}^\ell_i, \, \hat{y}^u_i] = g(\textbf{x}_i, \boldsymbol{\theta}_g)$.
Note that there is no assumption of $\hat{y}^\ell_i$ and $\hat{y}^u_i$  being symmetric with respect to the target estimate $\hat{y}_i$ produced by network $f(\cdot)$.
We describe its optimization procedure below.

We say that a training sample $\textbf{x}_i \in \textbf{X}^b$ is covered (i.e., we set $k_i = 1$) if both the predicted value $\hat{y}_i$ and the target observation $y_i$ fall within the estimated PI:
\begin{equation}
  k_i =
    \begin{cases}
      1, & \text{if $\hat{y}^\ell_i < \hat{y}_i < \, \hat{y}^u_i$ and $\hat{y}^\ell_i < y_i < \, \hat{y}^u_i$}\\
      0, & \text{otherwise}.
    \end{cases}
 \label{eq:k}
\end{equation}
\noindent Then, using $k_i$, we define the prediction interval coverage probability ($PICP$) for $\textbf{X}^b$ as the percent of covered samples with respect to the batch size $N$: $PICP = \sum_{i=1}^N k_i / N$.

The HQ principle suggests that the width of the prediction intervals should be minimized as long as they capture the target observation value.
Thus, Pearce et al.~[\citenum{ICML-2018-PearceBZN}] considered the mean prediction interval width of captured points ($MPIW_{capt}$) as part of their loss function:
\begin{equation}
    MPIW_{capt} =  \frac{1}{\epsilon + \sum_{i}k_i} \sum_{i=1}^N (\hat{y}^u_i - \hat{y}^\ell_i) \, k_i,
 \label{eq:mpiw_capt}
\end{equation}
\noindent where $\epsilon$ is a small number used to avoid dividing by zero.
However, we argue that minimizing $MPIW_{capt}$ does not imply that the width of the PIs generated for the non-captured samples will not decrease along with the width of the PIs generated for the captured samples\footnote{
We provide a toy example demonstrating this behavior in the following link 
\url{https://github.com/NISL-MSU/PredictionIntervals/tree/master/src/PredictionIntervals/models/QD_toy_example.ipynb}}. 

Furthermore, consider the case where none of the samples are captured by the PIs, as likely happens at the beginning of the training. Then, the penalty is minimum (i.e., $MPIW_{capt}=0$).
Hence, the calculated gradients of the loss function will force the weights of the NN to remain in the state where $\forall i, \; k_i = 0$, which contradicts the goal of maximizing $PICP$.

Instead of minimizing $MPIW_{capt}$ directly, we let
\begin{equation}
    PI_{pen} = \frac{1}{N} \sum_{i=1}^{N} (|\hat{y}^u_i - y_i| + |y_i - \hat{y}^\ell_i|),
    \label{eq:MPIW_pen}
\end{equation}
which we minimize instead. This function quantifies the width of the PI as the sum of the distance between the upper bound and the target and the distance between the lower bound and the target.
We argue that $PI_{pen}$ is more suitable than $MPIW_{capt}$ given that it forces $\hat{y}^u_i$, $y_i$, and $\hat{y}^\ell_i$ to be closer together.
For example, suppose that the following case is observed during the first training epoch: $y_i=24$, $\hat{y}_i= 25$, $\hat{y}^u_i=0.2$, and $\hat{y}^\ell_i=0.1$.
Then $MPIW_{capt} = 0$ given that the target is not covered by the PI, while $PI_{pen} = 47.7$.
As a result, $PI_{pen}$ will penalize this state while $MPIW_{capt}$ will not.
Thus, we define our first optimization objective as:
\[
    \min_{\boldsymbol{\theta}_g} \ \mathcal{L}_1 = \min_{\boldsymbol{\theta}_g} \ PI_{pen}.
\]

However, minimizing $\mathcal{L}_1$ is not enough to ensure the integrity of the PIs.
Their integrity is given by the conditions that the upper bound must be greater than the target and the target estimate ($\hat{y}^u_i > y_i$ and $\hat{y}^u_i > \hat{y}_i$) and that the target and the target estimate, in turn, must be greater than the lower bound ($y_i > \hat{y}^\ell_i$ and $\hat{y}_i > \hat{y}^\ell_i$).
Note that if the differences $(\hat{y}^u_i - y_i)$ and $(y_i - \hat{y}^\ell_i)$ are greater than the maximum estimation error within the training batch $\textbf{X}^b$ (i.e., $(\hat{y}^u_i - y_i) > \max_i |\hat{y}_i - y_i|$ and $(\hat{y}^u_i - y_i) > \max_i |\hat{y}_i - y_i|$, $\forall i \in [1, \dots, N]$), it is implied that all samples are covered ($k_i = 1, \, \forall i \in [1, \dots, N]$).

Motivated by this, we include an additional penalty function to ensure PI integrity and maximize the number of covered samples within the batch simultaneously.
Let us denote the mean differences between the PI bounds and the target estimates as $d_u = \sum_{i=1}^N(\hat{y}^u_i - y_i) / N$ and $d_\ell = \sum_{i=1}^N(y_i - \hat{y}^\ell_i) / N$.
Let $\xi = \max_i |\hat{y}_i - y_i|$
denote the maximum distance between a target estimate and its corresponding target value within the batch ($\xi > 0$).
From this, our penalty function is defined as:
\begin{equation}
    P = e^{\xi - d_u} + e^{\xi - d_\ell},
    \label{eq:C}
\end{equation}
Here, if the PI integrity is not met (i.e., $d_u < 0$ or $d_\ell < 0$) then their exponent magnitude becomes larger than $\xi$, producing a large penalty value.
Moreover, these terms encourage both $d_u$ and $d_\ell$ not only to be positive but also to be greater than $\xi$.
This implies that the distance between the target $y_i$ and any of its bounds will be larger than the maximum error within the batch, $\xi$, thus the target $y_i$ will lie within the PI.
From this, we define our second optimization objective as:
\[
    \min_{\boldsymbol{\theta}_g} \ \mathcal{L}_2 = \min_{\boldsymbol{\theta}_g} \ P.
\]
Then our proposed dual accuracy-quality-driven loss function is given by 
\begin{equation}
    Loss_{DualAQD} = \mathcal{L}_1 + \lambda \, \mathcal{L}_2,
    \label{eq:dualaqd}
\end{equation}
where $\lambda$ is a self-adaptive coefficient that controls the relative importance of $\mathcal{L}_1$ and $\mathcal{L}_2$.
Hence, our multi-objective optimization problem can be expressed as:
\[
    \boldsymbol{\theta}_g = \underset{\boldsymbol{\theta}_g}{\text{argmin}} \ Loss_{DualAQD}.
\]

For simplicity, we assume that $f(\cdot)$ and $g(\cdot)$ have $L$ layers and the same network architecture except for the output layer.
Network $f(\cdot)$ is trained first.
Then, weights $\boldsymbol{\theta}_g$ are initialized using weights $\boldsymbol{\theta}_f$ except for those of the last layer: $\boldsymbol{\theta}_g^{(0)}[1: L-1] = \boldsymbol{\theta}_f[1: L-1]$.
Note, that, in general, DualAQD can use different network architectures for $f(\cdot)$ and $g(\cdot)$.


\subsection{Batch Sorting} \label{sec:bs}
\begin{figure}[!t]
\centering
\includegraphics[width = \columnwidth]{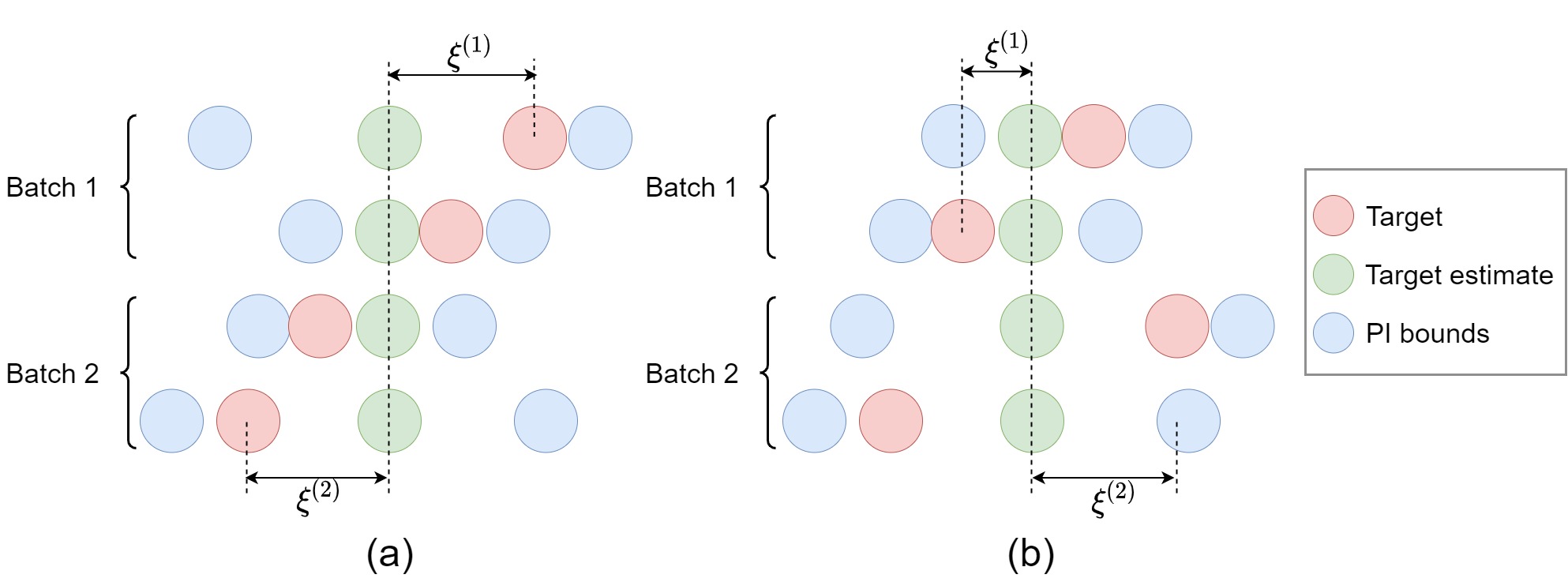}
\vspace{-4ex}
\caption{$\mathcal{L}_3$ penalty calculation, \textbf{(a)} without batch sorting;  \textbf{(b)} with batch sorting.}
\vspace{-1ex}
\label{fig:bs}
\end{figure}

The objective function $\mathcal{L}_2$ minimizes the term $P$ (Eq.~\ref{eq:C}), forcing the distance between the target estimate of a sample and its PI bounds to be larger than the maximum absolute error within its corresponding batch.
This term assumes there exists a similarity among the samples within a batch.
However, consider the case depicted in Fig.~\ref{fig:bs} where we show four samples that have been split randomly into two batches.
In Fig.~\ref{fig:bs}a, the PIs of the second and third samples already cover their observed targets.
Nevertheless, according to $\mathcal{L}_2$, these samples will yield high penalties because the distances between their target estimates and their PI bounds are less than $\xi^{(1)}$ and $\xi^{(2)}$, respectively, forcing their widths to increase unnecessarily.

For this reason, we propose a method called ``batch sorting", which consists of sorting the training samples with respect to their corresponding generated PI widths after each epoch.
By doing so, the batches will process samples with similar widths, avoiding unnecessary widening.
For example, in Fig.~\ref{fig:bs}b, the penalty terms are low given that $d_u^{(1)}, d_\ell^{(1)} > \xi^{(1)}$ and $d_u^{(2)}, d_\ell^{(2)} > \xi^{(2)}$. 
Note that, during testing, the PI generated for a given sample is independent of other samples and, as such, batch sorting becomes unnecessary during inference.

\subsection{Self-adaptive Coefficient $\lambda$} \label{sec:lambda}

The coefficient $\lambda$ of Eq.~\ref{eq:dualaqd} balances the two optimization objectives $\mathcal{L}_1$ and $\mathcal{L}_2$.
In this section, we propose that, instead of $\lambda$ being a tunable hyperparameter with a fixed value throughout training, it should be adapted throughout the learning process automatically.

Typically, the $PICP$ value improves as long as the $MPIW$ value increases;
however, extremely wide PIs are not useful. 
We usually aim to obtain PIs with a nominal probability coverage no greater than $(1- \alpha)$. 
A common value for the significance level $\alpha$ is $0.05$, in which case we say that we are 95\% confident that the target value will fall within the PI.

Let $PICP_{train}^{(t)}$ denote the $PICP$ value calculated on the training set $\textbf{X}_{train}$ after the $t$-th training epoch. 
If $PICP_{train}^{(t)}$ is below the confidence target $(1 - \alpha)$, more relative importance should be given to the objective $\mathcal{L}_2$ that enforces PI integrity (i.e., $\lambda$ should increase).
Likewise, if $PICP_{train}^{(t)}$ is higher than $(1 - \alpha)$, more relative importance should be given to the objective $\mathcal{L}_1$ that minimizes $MPIW$ (i.e., $\lambda$ should decrease).

We formalize this intuition by defining the cost $\mathcal{C}$ that quantifies the distance from $PICP_{train}^{(t)}$ to the confidence target $(1 - \alpha)$:
$
  \mathcal{C} = (1 - \alpha) - PICP_{train}^{(t)}.
$
Then, we propose to increase or decrease $\lambda$ proportionally to the cost function $\mathcal{C}$ after each training epoch as follows (see Algorithm~\ref{alg:lambda}):
\begin{equation}
    \lambda^{(t)} = \lambda^{(t - 1)} + \eta \cdot \mathcal{C},
    \label{eq:lambda}
\end{equation}
where $\lambda^{(t)}$ is the value of the coefficient $\lambda$ at the $t$-th iteration (we consider that $\lambda^{(0)}=1$), and $\eta$ is a tunable scale factor. 

Note that Algorithm~\ref{alg:lambda} takes as inputs the data $\textbf{X}_{train}$ and corresponding targets $Y_{train}$ as well as the trained prediction network $f$, the untrained network $g$, the significance level $\alpha$, and the scale factor $\eta$.
Function {\tt batchSorting}$(\textbf{X}_{train}, Y_{train}, widths^{(t-1)})$ returns a list of batches sorted according to the PI widths generated during the previous training epoch (see Sec.\ref{sec:bs}).
Function {\tt DualAQD}$(\lambda, y, \hat{y}, \hat{y}^u, \hat{y}^\ell)$ represents the DualAQD loss function (Eq.\ref{eq:dualaqd}) while {\tt update}($g, loss$) encompasses the conventional backpropagation and gradient descent processes used to update the weights of network $g$.
Furthermore, function {\tt metrics}$(\textbf{X}_{train}, Y_{train})$ passes $\textbf{X}_{train}$ through $g$ to generate the corresponding PIs and their widths, and to calculate compares the output to $Y_{train}$ to calculate the $PICP_{train}^{(t)}$ value using $Y_{train}$. 

\begin{algorithm} [t]
\scriptsize
\caption{DualAQD method}
\begin{algorithmic}[1]
\Function{TrainNNwithDualAQD}{$\textbf{X}_{train}, Y_{train}, f, g, \alpha, \eta$}
    \State $\lambda \leftarrow 1$
    \For { each $t\in$ range(1, $maxEpochs$) }
        \If {$t > 1$}
            \State $Batches \leftarrow batchSorting(\textbf{X}_{train}, Y_{train}, widths)$
        \Else 
            \State $Batches \leftarrow shuffle(\textbf{X}_{train}, Y_{train})$
        \EndIf
        \For { each $batch \in Batches$ }
            \State $\textbf{x}, y \leftarrow batch$
            \State $\hat{y} \leftarrow f(\textbf{x})$ 
            \State $\hat{y}^u, \hat{y}^\ell \leftarrow g(\textbf{x})$ 
            \State $loss \leftarrow DualAQD(\lambda, y, \hat{y}, \hat{y}^u, \hat{y}^\ell)$
            \State $update(g, loss)$
        \EndFor
        \State $PICP_{train}^{(t)}, widths^{(t)} \leftarrow metrics(\textbf{X}_{train}, Y_{train})$
        \State // Update coefficient $\lambda$
        \State $\mathcal{C} \leftarrow ((1 - \alpha) - PICP_{train}^{(t)})$
        \State $\lambda = \lambda + \eta \cdot \mathcal{C}$
    \EndFor
    \State \Return $g$
\EndFunction
\end{algorithmic}
\label{alg:lambda}
\end{algorithm}


\subsection{Parameter and Hyperparameter Selection} \label{sec:hyp}

We train a neural network on the training set $\textbf{X}_{train}$ during $T$ epochs using $Loss_{DualAQD}$ as the loss function.
After the $t$-th training epoch, we calculate the performance metrics $z_t = \{ PICP_{val}^{(t)}, MPIW_{val}^{(t)} \}$ on the validation set $\textbf{X}_{val}$.
Thus, we consider that the set of optimal weights of the network, $\theta_g$, will be those that maximize performance on the validation set.
The remaining question is what are the criteria to compare two solutions $z_i$ and $z_j$.

Taking this criterion into account, we consider that a solution $z_{i}$ dominates another solution $z_{j}$ ($z_{i} \preceq z_{j}$) if: 
\begin{itemize}
    \item $PICP_{val}^{(i)} > PICP_{val}^{(j)}$ and $PICP_{val}^{(i)} \leq (1 - \alpha)$.
    \item $PICP_{val}^{(i)} == PICP_{val}^{(j)} < (1 - \alpha)$ and $MPIW_{val}^{(i)} < MPIW_{val}^{(j)}$
    \item $PICP_{val}^{(i)} \geq (1 - \alpha)$ and $MPIW_{val}^{(i)} < MPIW_{val}^{(j)}$
\end{itemize}
In other words, if $\alpha = 0.05$, we seek a solution whose $PICP_{val}$ value is at least 95\%.
After exceeding this value, a solution $z_i$ is said to dominate another solution $z_j$ only if it produces narrower PIs.

We use a grid search to tune the hyperparameter $\eta$ for training (Eq.~\ref{eq:lambda}).
For each value, we train a NN using 10-fold cross-validation and calculate the average performance metrics on the validation sets.
Then, the hyperparameters are selected using the dominance criteria explained above.


\subsection{PI Aggregation Using MC-Dropout}  \label{sec:MC}

In Sec.~\ref{sec:intro}, we explained that both the model uncertainty ($\sigma^2_{model}$) and the data noise variance ($\sigma^2_{noise}$) have to be taken into account when generating PIs.
A model trained using $Loss_{DualAQD}$ generates PI estimates based on the training data; that is, it accounts for $\sigma^2_{noise}$. 
However, we still need to quantify the uncertainty of those estimates due to $\sigma^2_{model}$.

Unlike previous work that used explicit NN ensembles to quantify $\sigma^2_{model}$ [\citenum{MVEEns,ICML-2018-PearceBZN}], we propose to use a Monte Carlo-based approach.
Specifically, we use MC-Dropout~\cite{dropout}, which consists of using dropout layers that ignore each neuron of the network according to some probability or dropout rate.
Then, during each forward pass with active dropout layers, a slightly different network architecture is used and, as a result, a slightly different prediction is obtained.
According to Gal and Ghahramani~[\citenum{pmlr-v48-gal16}], this process can be interpreted as a Bayesian approximation of the Gaussian process.

Our approach consists of using $M$ forward passes through the network with active dropout layers.
Given an input $\textbf{x}_i$, the estimates $\hat{y}_{i}^{(m)}$, $\hat{y}_i^{u(m)}$, and $\hat{y}_i^{\ell(m)}$ are obtained at the $m$-th iteration. 
Hence, the expected target estimate $\bar{y}_{i}$, the expected upper bound $\bar{y}^u_i$, and the expected lower bound $\bar{y}^\ell_i$ are calculated as:
$
    \bar{y}_{i} = \frac{1}{M} \sum_{m=1}^M \hat{y}_{i}^{(m)},   \bar{y}^u_i = \frac{1}{M} \sum_{m=1}^M \hat{y}_i^{u(m)},
    \bar{y}^\ell_i = \frac{1}{M} \sum_{m=1}^M \hat{y}_i^{\ell(m)}.
$

\subsection{Comparison to QD-Ens and QD+} \label{sec:comparison}

Here we consider the differences between our method (DualAQD) and the two methods QD-Ens~[\citenum{ICML-2018-PearceBZN}] and QD+~[\citenum{pmlr-v124-saleh-salem20a}].
For reference, we include the loss functions used by QD-Ens and QD+:

\begin{equation*}
\begin{aligned}
    Loss_{QD} = \, &MPIW_{capt}  + \\
    &\delta \, \frac{N}{\alpha (1 - \alpha)} \max(0, (1 - \alpha) - PICP) ^ 2.\\
     Loss_{QD+} =\, & (1 - \lambda_1) (1 - \lambda_2) MPIW_{capt}  + \\
     &\lambda_1 (1 - \lambda_2) \max(0, (1 - \alpha) - PICP) ^ 2 +\\
     &\lambda_2 \, MSE_{est} + \\
     & \frac{\xi}{N} \sum_{i=1}^N \left[ \max(0, (\hat{y}^u_i - \hat{y}_i) + \max(0, (\hat{y}_i - \hat{y}^\ell_i) \right],
\end{aligned}
    \label{eq:QD}
\end{equation*}

\noindent where $\delta$, $\lambda_1$, $\lambda_2$, and $\xi$ are hyperparameters used by QD-Ens and QD+ to balance the learning objectives. 
The differences compared to our method are listed in order of importance from highest to lowest as follows:

\begin{itemize}[leftmargin=*]
    \item QD-Ens and QD+ use objective functions that maximize $PICP$ directly aiming to a goal of $(1 - \alpha)$ at the batch level.
    We maximize $PICP$ indirectly through $\mathcal{L}_2$, which encourages the model to produce PIs that cover as many training points as possible.
    This is achieved by producing PIs whose widths are larger than the maximum absolute error within each training batch.
    Then the optimal weights of the network are selected as those that produce a coverage probability on the validation set of at least $(1 - \alpha)$.
    \item Note that $PICP$ is not directly differentiable as it involves counting the number of samples that lay within the predicted PIs.
    However, QD-Ens and QD+ force its differentiation by including a sigmoid operation and a softening factor (i.e., an additional hyperparameter). 
    On the other hand, the loss functions of DualAQD are already differentiable.
    \item Our objective $\mathcal{L}_1$ minimizes $PI_{pen}$, which is a more suitable penalty function than $MPIW_{capt}$ (cf. Sec~\ref{sec:loss}). 
    \item Our objective $\mathcal{L}_2$ maximizes $PICP$ and ensures PI integrity simultaneously.
    QD+ uses a truncated linear constraint and a separate function to maximize $PICP$.
    \item NN-based PI generation methods aim to balance three objectives: (1) accurate target prediction, (2) generation of narrow PIs, and (3) high coverage probability.
    QD-Ens uses a single coefficient $\delta$ within its loss function that balances objectives (2) and (3) and does not optimize objective (1) explicitly, while QD+ uses three coefficients $\lambda_1$, $\lambda_2$, and $\xi$ to balance the three objectives.
    All of the coefficients are tunable hyperparameters.
    Our loss function, $Loss_{DualAQD}$, uses a balancing coefficient whose value is not fixed but is adapted throughout the training process using a single hyperparameter (i.e., the scale factor $\eta$).
    \item Our approach uses two companion NNs $f(\cdot)$ and $g(\cdot)$ that optimize objective (1) and objectives (2) and (3), respectively, to avoid the trade-off between them.
    Conversely, the other approaches optimize a single NN architecture.
    \item We use MC-Dropout to estimate the model uncertainty. By doing so, we need to train only a single model instead of using an explicit ensemble of models, as in QD-Ens and QD+. Also,
    QD+ requires fitting a split normal density function~[\citenum{Wallis2014TheTN}] for each data point to aggregate the PIs produced by the ensemble, thus increasing the complexity of the learning process.
\end{itemize}

\section{Experiments}  \label{sec:results}

\subsection{Experiments with Synthetic Data} \label{sec:synth}

Previous approaches have been tested on datasets with similar uncertainty levels across all their samples, or on synthetic datasets with a single region of low uncertainty surrounded by a gradual increase of noise.
This is a limitation as it does not allow testing the ability of the PI's to adapt to rapid changes of uncertainty within the data.
Therefore, we test all of the methods on a more challenging synthetic dataset with more fluctuations and extreme levels of uncertainty. 
The code is available at \url{https://github.com/NISL-MSU/PredictionIntervals}.

We created a synthetic dataset with varying PI widths that consists of a sinusoid with Gaussian noise.
Specifically, the dataset contains 1000 points generated using the equation $y(x) = 5 \, \text{cos}(x) + 10 + \epsilon$, where $x \in [-5, 5]$ and $\epsilon$ is Gaussian noise whose magnitude depends on $x$: $\epsilon = (2 \, \text{cos}(1.2 \, x) + 2) \, v$ where $v \sim \mathcal{N}(0, 1)$.  
For these experiments, we trained a feed-forward neural network with two hidden layers, each with 100 nodes with ReLU activation.
A $5\times2$-fold cross-validation design was used to train and evaluate all networks.

Knowing the probability distribution of the noise at each position $x$ allows us to calculate the ideal 95\% PIs ($\alpha = 0.05$), $[y^u, y^\ell]$, as follows:
\[
y^u(x) = y(x) + 1.96 \, \epsilon, \text{  and  } 
y^\ell(x) = y(x) - 1.96 \, \epsilon, 
\]
where $1.96$ is the approximate value of the 95\% confidence interval of the normal distribution.
Therefore, we define a new metric we called $PI_\delta$ that sums the absolute differences between the estimated bounds and the ideal 95\% bounds for all the samples within a set $\textbf{X}$:
\[
PI_\delta = \frac{1}{\left| \textbf{X} \right|} \sum_{x \in \textbf{X}} \left( |y^u(x) - \hat{y}^u(x)| + | y^\ell(x) - \hat{y}^\ell(x)| \right).
\]

We compared the performance of DualAQD using batch sorting and without using batch sorting (denoted as ``DualAQD\_noBS" in Table~\ref{tab:synth}).
All networks were trained using a fixed mini-batch size of 16 and the Adadelta optimizer.
Table~\ref{tab:synth} gives the average performance for the metrics calculated on the validation sets, $MSE_{val}$, $MPIW_{val}$, $PICP_{val}$, and ${PI_\delta}_{val}$, and corresponding standard deviations.

We also compared our DualAQD PI generation methodology to three other NN-based methods: QD+~[\citenum{pmlr-v124-saleh-salem20a}], QD-Ens~[\citenum{ICML-2018-PearceBZN}], and a PI generation method based on MC-Dropout alone~[\citenum{MCPI}] (denoted MC-Dropout-PI).
For the sake of consistency and fairness, we used the same configuration (i.e., network architecture, optimizer, and batch size) for all the networks trained in our experiments.
In our preliminary experiments, for the case of QD+, QD-Ens, and MC-Dropout-PI, we found that batch sorting either helped to improve their performance or there was no significant change. 
Thus, for the sake of fairness and consistency, we decided to use batch sorting for all compared methods. 
In addition, we tested Dropout rates between $0.1$ and $0.5$. 
The obtained results did not indicate a statistically significant difference; thus, we used a Dropout rate of 0.1 for all networks and datasets.

\begin{figure}[!t]
\centering
\includegraphics[width = 8cm]{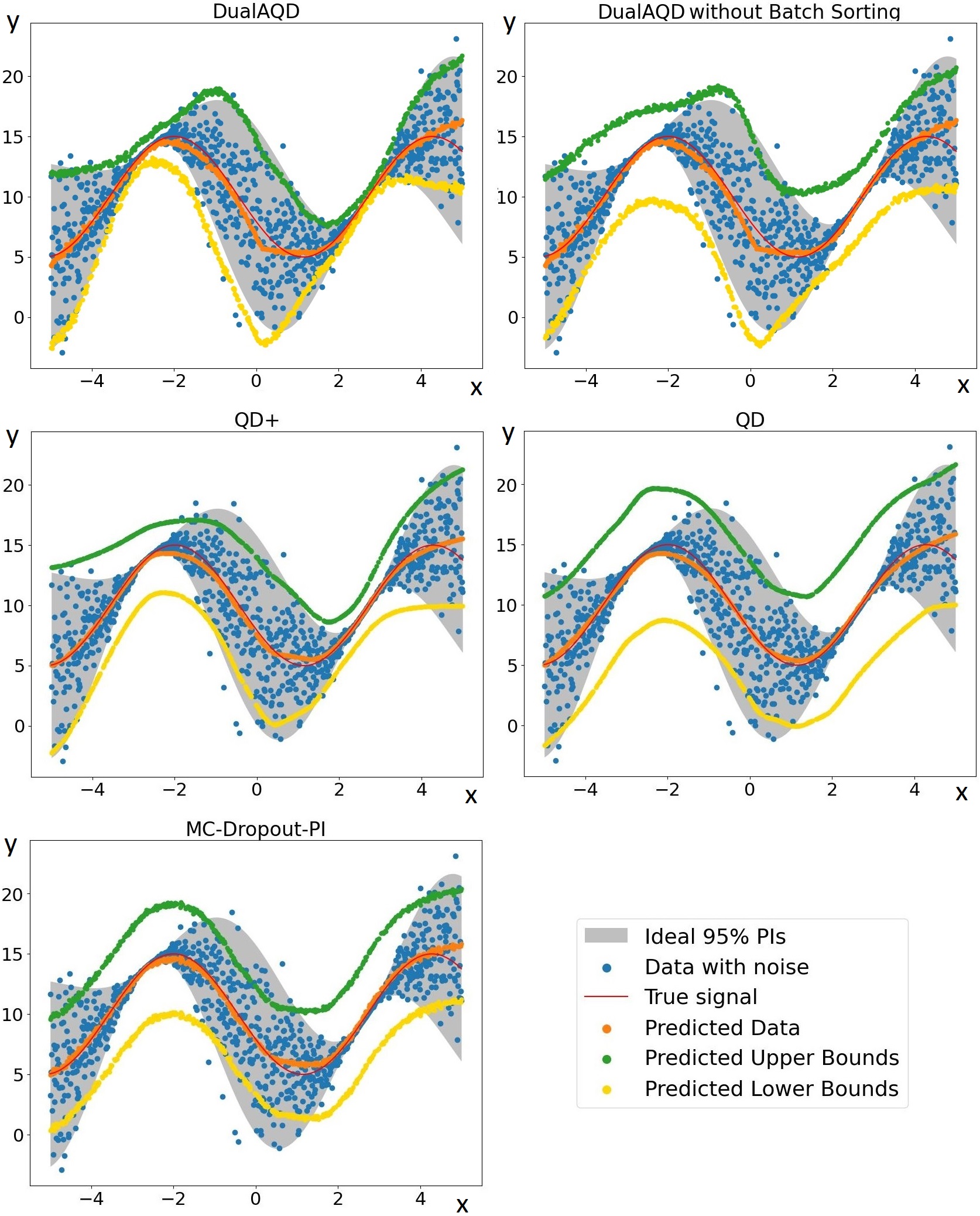}
\caption{Performance of PI generation methods on the synthetic dataset.}
\vspace{-2ex}
\label{fig:synth}
\end{figure}

Note that the only difference between the network architecture used by the four methods is that QD+ requires three outputs, QD-Ens requires two (i.e., the lower and upper bounds), and MC-Dropout-PI requires one.
For DualAQD and MC-Dropout-PI, we used $F=100$ forward passes with active dropout layers.
For QD+ and QD-Ens, we used an ensemble of five networks and a grid search to choose the hyperparameter values.
Fig.~\ref{fig:synth} shows the PIs generated by the four methods from the first validation set together with the ideal 95\% PIs.

\begin{table}[!t]
\begin{center}
\caption{PI metrics \textbf{$MSE_{val}$}, \textbf{$MPIW_{val}$}, \textbf{$PICP_{val}$}, and \textbf{${PI_\delta}_{val}$} evaluated on the synthetic dataset using $5 \times 2$ cross-validation.}
\label{tab:synth}
\vspace{-2ex}
\resizebox{\columnwidth}{!}{
\begin{tabular}{|c|c|c|c|c|}
\hline
\textbf{Method} & \textbf{$MSE_{val}$} & \textbf{$MPIW_{val}$} & \textbf{$PICP_{val} (\%)$} & \textbf{${PI_\delta}_{val}$} \\ \hline
DualAQD & 5.27 $\pm$ 0.27 & 7.30 $\pm$ 0.29 & 95.5 $\pm$ 0.48 & 1.52 $\pm$ 0.13 \\ \hline
DualAQD\_noBS & 5.27 $\pm$ 0.27 & 9.16 $\pm$ 0.35 & 96.3 $\pm$ 0.77& 3.08 $\pm$ 0.19 \\ \hline
QD+ & 5.28 $\pm$ 0.29 & 8.56 $\pm$ 0.14 & 95.5 $\pm$ 0.31& 3.12 $\pm$ 0.24\\ \hline
QD-Ens & 5.31 $\pm$ 0.26 & 10.17 $\pm$ 0.79 & 94.0 $\pm$ 1.57& 4.88 $\pm$ 0.17 \\ \hline
MC-Dropout-PI & 5.22 $\pm$ 0.30 & 9.31 $\pm$ 0.27 & 93.3 $\pm$ 0.63& 5.04 $\pm$ 0.08 \\ \hline
\end{tabular}
}
\vspace{-2ex}
\end{center}
\end{table}

\subsection{Benchmarking Experiments} \label{sec:bench_exp}

We experimented with eight open-access datasets from the UC Irvine Machine Learning Repository~[\citenum{Dua:2019}].
Note that even though our experiments use scalar and 2-D regression tasks (Sec.~\ref{sec:Yield}), our proposed method can be extended to other tasks such as classification. 
For each dataset, we used a feed-forward neural network whose architecture was the same as that described in Sec.~\ref{sec:synth}.
We used 10-fold cross-validation to train and evaluate all networks.
Table~\ref{tab:bench} gives the average performance for the metrics calculated on the validation sets, $MSE_{val}$, $MPIW_{val}$, and $PICP_{val}$, and corresponding standard deviations.
We applied $z$-score normalization (mean equal to 0 and standard deviation equal to 1) to each feature in the training set while the exact same scaling was applied to the features in the validation and test sets.
Likewise, min-max normalization was applied to the response variable; however, Table~\ref{tab:bench} shows the results after re-scaling to the original scale.
Similar to Sec.~\ref{sec:synth}, all networks were trained using a fixed mini-batch size of 16, except for the Protein and Year datasets that used a mini-batch size of 512 due to their large size. 

\begin{figure*}
    \centering
    \includegraphics[width = \textwidth]{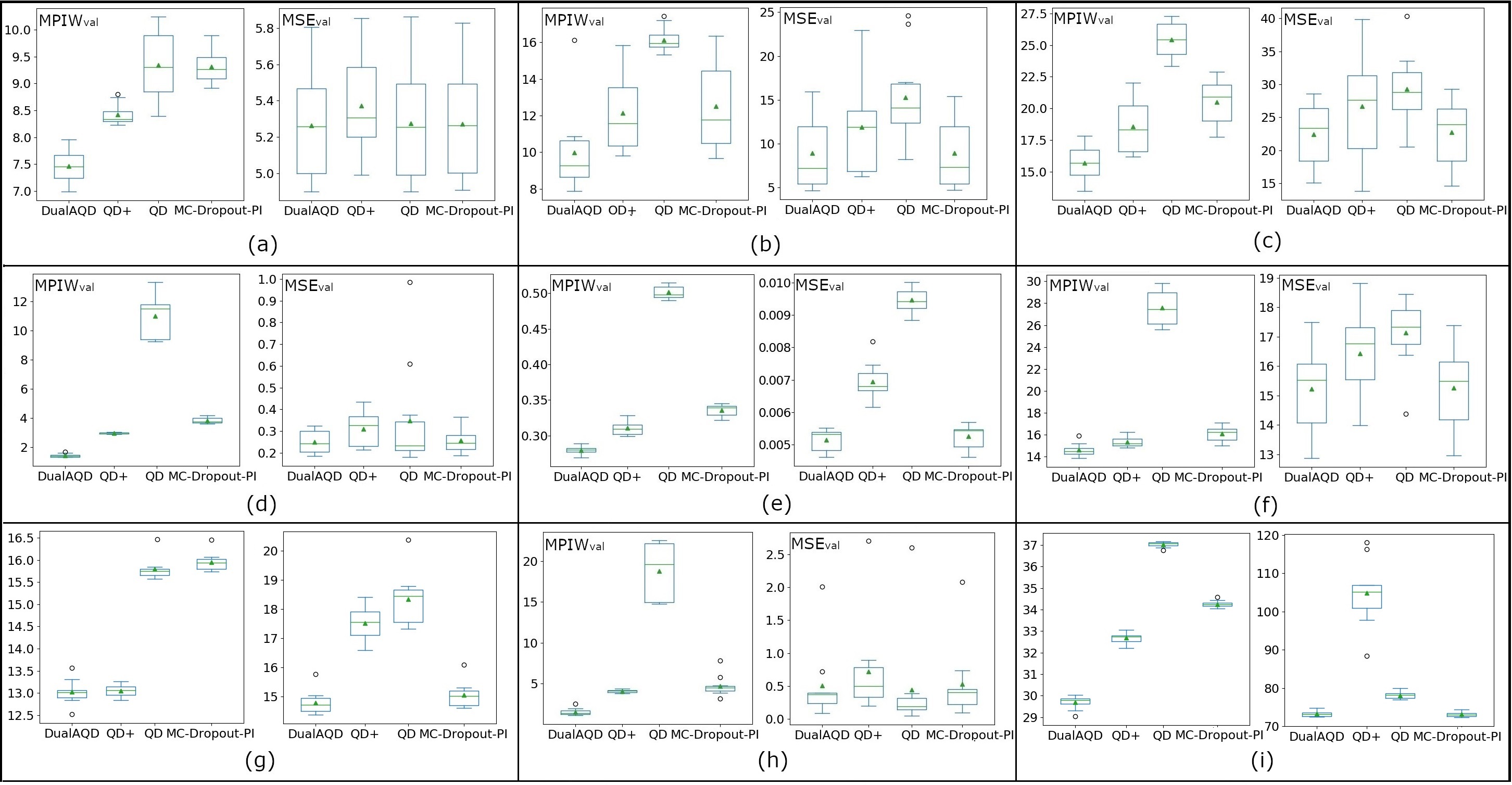}
    \caption{Box plots of the $MPIW_{val}$ and $MSE_{val}$ scores of DualAQD, QD+, QD-Ens, and MC-Dropout-PI PI generation methods on the synthetic and benchmarking datasets: \textbf{(a)} Synthetic. \textbf{(b)} Boston. \textbf{(c)} Concrete. \textbf{(d)} Energy. \textbf{(e)} Kin8nm. \textbf{(f)} Power. \textbf{(g)} Protein. \textbf{(h)} Yacht. \textbf{(i)} Year.
    \vspace{-2ex}}
    \label{fig:box_plots}
\end{figure*}

The bold entries in Table~\ref{tab:bench} indicate the method that achieved the lowest average $MPIW_{val}$ value and that its difference with respect to the values obtained by the other methods is statistically significant according to a paired $t$-test performed at the 0.05 significance level. 
The results obtained by DualAQD were significantly narrower than the compared methods while having similar $MSE_{val}$ and $PICP_{val}$ of at least 95\%.
Furthermore, Fig.~\ref{fig:box_plots} depicts the distribution of the scores achieved by all the compared methods on all the datasets, where the line through the center of each box indicates the median F1 score, the edges of the boxes are the 25th and 75th percentiles, whiskers extend to the maximum and minimum points (not counting outliers), and
outlier points are those past the end of the whiskers (i.e., those points greater than $1.5\times IQR$ plus the third quartile or less than $1.5\times IQR$ minus the first quartile, where $IQR$ is the inter-quartile range).

\begin{figure}
    \centering
    \includegraphics[width = 5.2cm]{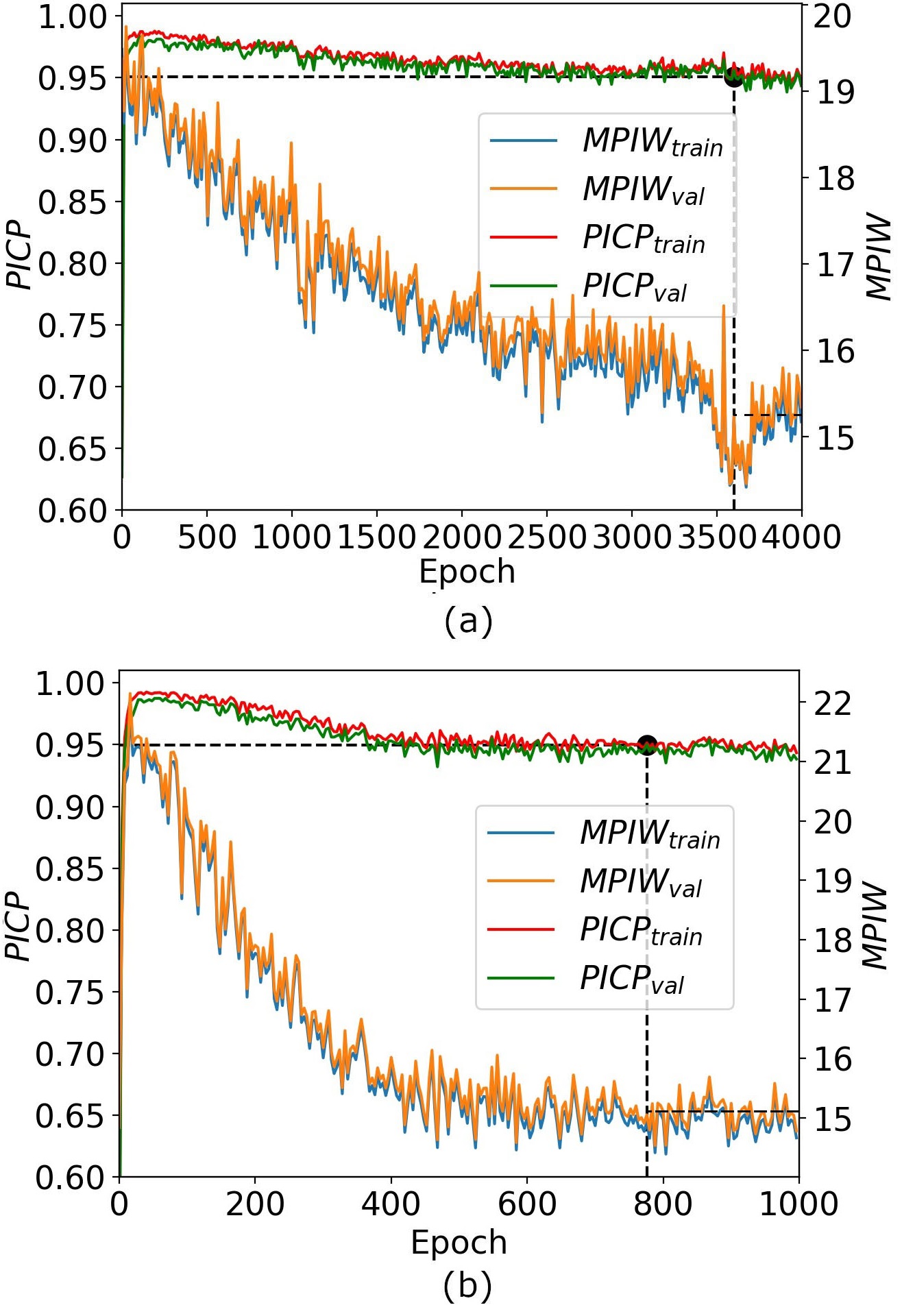}
    \vspace{-2ex}
    \caption{$MPIW$ and $PICP$ learning curves obtained for the Power dataset using DualAQD. \textbf{(a)} $\eta=0.01$. \textbf{(b)} $\eta=0.1$.}
    \label{fig:Power_curves}
\end{figure}

Note that even though QD-Ens uses only one hyperparameter (see Sec.~\ref{sec:comparison}), it is more sensitive to small changes.
For example, a hyperparameter value of $\delta = 0.021$ yielded poor PIs with $PICP_{val} < 40\%$ while a value of $\delta = 0.02105$ yielded too wide PIs with $PICP_{val} < 100\%$.
For this reason, the hyperparameter $\delta$ of the QD-Ens approach was chosen manually while the scale factor $\eta$ of DualAQD was chosen using a grid search with values $\{0.001, 0.005, 0.01, 0.05, 0.1\}$.
Fig.~\ref{fig:Power_curves} 
shows the difference between the learning curves obtained during one iteration of the cross-validation for the Power dataset using two different $\eta$ values (i.e., $\eta = 0.01$ and $\eta = 0.1$).
The dashed lines indicate the training epoch at which the optimal weights $\boldsymbol{\theta}_g$ were selected according to the dominance criteria explained in Sec.~\ref{sec:hyp}. 
On the other hand, the hyperparameters $\lambda_1$ and $\lambda_2$ of QD+ were chosen using a random search since it requires significantly higher training and execution time. 

\begin{table}[!t]
\begin{center}
\caption{PI metrics \textbf{$MSE_{val}$}, \textbf{$MPIW_{val}$}, and \textbf{$PICP_{val}$} evaluated on the benchmark datasets using $10$-fold cross-validation.}
\label{tab:bench}
\vspace{-1ex}
\def\arraystretch{1.2}%
\resizebox{\columnwidth}{!}{
\begin{tabular}{|c|c|c|c|c|c|}
\hline
\textbf{Dataset} & \textbf{Metric} & \textbf{DualAQD} & \textbf{QD+} & \textbf{QD-Ens} & \textbf{\begin{tabular}[c]{@{}c@{}}MC-\\ Dropout-PI\end{tabular}}  \\ \Xhline{3\arrayrulewidth}
\multirow{3}{*}{Boston}
 & $MPIW_{val}$  & \textbf{9.99$\pm$2.26} & 12.14$\pm$2.05 & 16.13$\pm$0.67 & 12.52$\pm$2.28  \\ \cline{2-6} 
 & $MSE_{val}$ & 8.91$\pm$3.90 & 11.91$\pm$5.24& 15.29$\pm$5.07& 8.94$\pm$3.87   \\ \cline{2-6} 
 & $PICP_{val} (\%)$ & 95.0$\pm$1.6 & 95.6$\pm$1.9  & 97.2$\pm$1.3 & 96.0$\pm$0.9 \\ \Xhline{3\arrayrulewidth}
\multirow{3}{*}{Concrete} 
 & $MPIW_{val}$ & \textbf{15.72$\pm$1.42} & 18.57$\pm$2.06 & 25.42$\pm$1.30 & 20.52$\pm$1.74 \\ \cline{2-6} 
 & $MSE_{val}$ & 22.45$\pm$4.79 & 26.65$\pm$8.02 & 29.30$\pm$5.25 & 22.71$\pm$4.96   \\ \cline{2-6} 
 & $PICP_{val} (\%)$ & 95.2$\pm$0.5 & 95.2$\pm$1.3 & 97.9$\pm$1.6 & 95.7$\pm$1.2 \\ \Xhline{3\arrayrulewidth}
\multirow{3}{*}{Energy} 
 & $MPIW_{val}$ & \textbf{1.41$\pm$0.12}& 2.94$\pm$0.05 & 10.99$\pm$1.47 & 3.81$\pm$0.21 \\ \cline{2-6} 
 & $MSE_{val}$ & 0.25$\pm$0.05& 0.31$\pm$0.08 & 0.35$\pm$0.25 & 0.26$\pm$0.05   \\ \cline{2-6} 
 & $PICP_{val} (\%)$ & 96.5$\pm$0.6& 99.0$\pm$1.0 & 100.0$\pm$0.0 & 99.5$\pm$0.6 \\ \Xhline{3\arrayrulewidth}
\multirow{3}{*}{Kin8nm} 
 & $MPIW_{val}$ & \textbf{0.280$\pm$0.01}& 0.311$\pm$0.01 & 0.502$\pm$0.01 & 0.336$\pm$0.01 \\ \cline{2-6} 
 & $MSE_{val}$ & 0.005$\pm$0.00& 0.007$\pm$0.00 & 0.009$\pm$0.00 & 0.005$\pm$0.00   \\ \cline{2-6} 
 & $PICP_{val} (\%)$ & 95.1$\pm$0.1& 96.6$\pm$0.4 & 98.5$\pm$0.3 & 97.5$\pm$0.4 \\ \Xhline{3\arrayrulewidth}
\multirow{3}{*}{Power} 
 & $MPIW_{val}$ & \textbf{14.60$\pm$0.35}& 15.31$\pm$0.44  & 27.57$\pm$1.54  & 16.08$\pm$0.63 \\ \cline{2-6} 
 & $MSE_{val}$ & 15.23$\pm$1.34& 16.43$\pm$1.34  & 17.14$\pm$1.11 & 15.26$\pm$1.31 \\ \cline{2-6} 
 & $PICP_{val} (\%)$ & 95.2$\pm$0.1& 95.7$\pm$0.3   & 99.6$\pm$0.2 & 96.4$\pm$0.5 \\ \Xhline{3\arrayrulewidth}
\multirow{3}{*}{Protein} 
 & $MPIW_{val}$ & \textbf{13.02$\pm$0.26} &  \textbf{13.05$\pm$0.14}  & 15.79$\pm$0.24  & 15.95$\pm$0.20 \\ \cline{2-6} 
 & $MSE_{val}$ & 14.79$\pm$0.40& 17.51$\pm$0.59  & 18.35$\pm$0.87 & 15.05$\pm$0.42 \\ \cline{2-6} 
 & $PICP_{val} (\%)$ & 95.0$\pm$0.1& 95.4$\pm$0.4   & 95.1$\pm$0.5 & 94.8$\pm$0.1 \\ \Xhline{3\arrayrulewidth}
\multirow{3}{*}{Yacht} 
 & $MPIW_{val}$ & \textbf{1.56$\pm$0.42} &4.10$\pm$0.17 & 10.99$\pm$1.47 & 4.74$\pm$1.20 \\ \cline{2-6} 
 & $MSE_{val}$ & 0.51$\pm$0.53 & 0.72$\pm$0.70 & 0.35$\pm$0.25 & 0.53$\pm$0.54   \\ \cline{2-6} 
 & $PICP_{val} (\%)$ & 97.1$\pm$0.9 & 98.4$\pm$2.2 & 100.0$\pm$0.0 & 100.0$\pm$0.0 \\ \Xhline{3\arrayrulewidth}
\multirow{3}{*}{Year} 
 & $MPIW_{val}$ & \textbf{29.68$\pm$0.29} &32.68$\pm$0.25 & 37.03$\pm$0.13 & 34.25$\pm$0.16\\ \cline{2-6} 
 & $MSE_{val}$ & 73.26$\pm$0.76 & 104.8$\pm$8.1 & 78.12$\pm$0.87 &  73.13$\pm$0.69   \\ \cline{2-6} 
 & $PICP_{val} (\%)$ & 95.1$\pm$0.1 & 95.4$\pm$0.9 & 37.03$\pm$0.1 & 93.82$\pm$0.0 \\ \Xhline{3\arrayrulewidth}
\end{tabular}
}
\end{center}
\end{table}

\subsection{Prediction Intervals for Crop Yield Prediction} \label{sec:Yield}

We assert our approach is general in applicability.
To test this assertion, we decided to experiment with a difficult, real-world application of 2D regression using spatially correlated data to convey the usefulness of our method.
Specifically, we focused on the crop yield prediction problem, which has an important impact on society and is one of the main tasks of precision agriculture.
Accurate and reliable crop yield prediction, along with careful uncertainty management strategies, enables farmers to make informed management decisions, such as determining the nitrogen fertilizer rates needed in specific regions of their fields to maximize profit while minimizing environmental impact~[\citenum{hegedus}].

We use an early-yield prediction dataset of winter wheat we curated and presented in a previous work \cite{Morales_2023}.
The early-yield prediction is posed as a regression problem where the explanatory variables are represented by a set of eight features obtained during the growing season (March).
These features consist of nitrogen rate applied, precipitation, slope, elevation, topographic position index (TPI), aspect, and two backscattering coefficients obtained from synthetic aperture radar (SAR) images from Sentinel-I.
The response variable corresponds to the yield value in bushels per acre (bu/ac), measured during the harvest season (August).
In other words, the data acquired in March is used to predict crop yield values in August of the same year.

The yield prediction problem requires two-dimensional (2D) inputs and 2D outputs.
As such, it can be viewed as a 2D regression task.
To tackle this problem, we trained a CNN using the Hyper3DNetReg 
3D-2D network, architecture we presented in \cite{Morales_2023}, which was specifically designed to predict the yield values of small spatial neighborhoods of a field simultaneously.
We then modified this architecture to produce three output patches of $5 \times 5$ pixels (i.e., the estimated yield patch and two patches containing the upper and lower bounds of each pixel, respectively) instead of one.

For our experiments, we used data collected from three winter wheat fields, which we refer to as ``A,'' ``B,'' and ``C'', respectively. 
Three crop years of data were collected for each field.
The information from the first two years was used to create the training and validation sets (90\% of the data is used for
training and 10\% for validation).
The four methods, AQD, QD+, QD-Ens, and MC-Dropout-PI, were compared using the results from the test set of each field, which consists of data from the last observed year and whose ground-truth yield map is denoted as $Y$.
The test set was used to generate a predicted yield map of the entire field, $\hat{Y}$, and its corresponding lower and upper bounds, $\hat{Y}_L$ and $\hat{Y}_U$, respectively.

\begin{figure}[!t]
\centering
\includegraphics[width = \columnwidth]{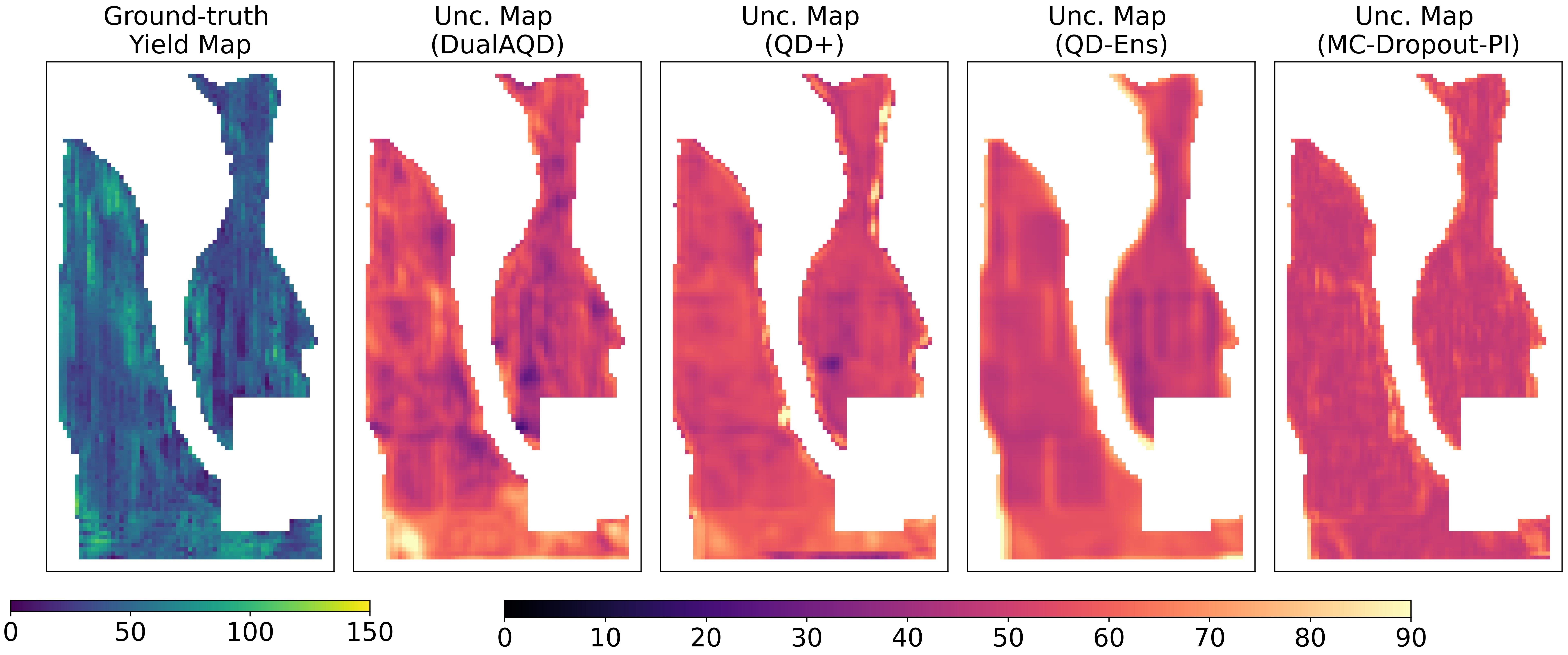}
\caption{Uncertainty maps comparison for field A. }
\label{fig:yield}
\end{figure}

Fig.~\ref{fig:yield} shows the ground-truth yield map for field ``A'' (darker colors represent lower yield values) along with the uncertainty maps obtained by the four compared methods and their corresponding $PICP$ and $MPIW$ values.
Field ``A'' is used as a representative field for presenting our results, since we obtained similar results on the other fields.
Here, we define the uncertainty map $U = \hat{Y}^u - \hat{Y}^\ell$ as a map that contains the PI width of each point of the field (darker colors represent lower PI width and thus lower uncertainty).
That is, the wider the PI of a given point, the more uncertain its yield prediction.

We used four metrics to assess the behavior of the four methods (Table~\ref{tab:comp1}). 
First, we calculated the root mean square error ($RMSE_{test}$) between the ground-truth yield map $Y$ and the estimated yield map $\hat{Y}$. 
Then, we considered the mean prediction interval width ($MPIW_{test}$) and prediction interval probability coverage ($PICP_{test}$). 
Note that $k$-fold or $k\times 2$ cross-validation cannot be used in this experimental setting.
Thus, to help us explain the advantages of our method over the others in the context of the HQ principle, we introduce a new metric that summarizes the $MPIW_{test}$ and $PICP_{test}$ metrics shown in Table~\ref{tab:comp1}.
Let $\overline{MPIW}_{test}$ represent the mean PI width after min-max normalization using as upper bound the maximum $MPIW_{test}$ value among the four methods in each field.  
Let $\mu_\omega$ denote the weighted geometric mean between $\overline{MPIW}_{test}$ and ($1 - PICP_{test})$ (i.e., the complement of the PI coverage probability) with $\omega \in [0, 1]$ being the relative importance between both terms. Then
\[
    \mu_\omega = (\overline{MPIW}_{test})^\omega (1 - PICP_{test})^{(1 - \omega)}.
\]

\begin{figure}[!t]
\centering
\includegraphics[width = \columnwidth]{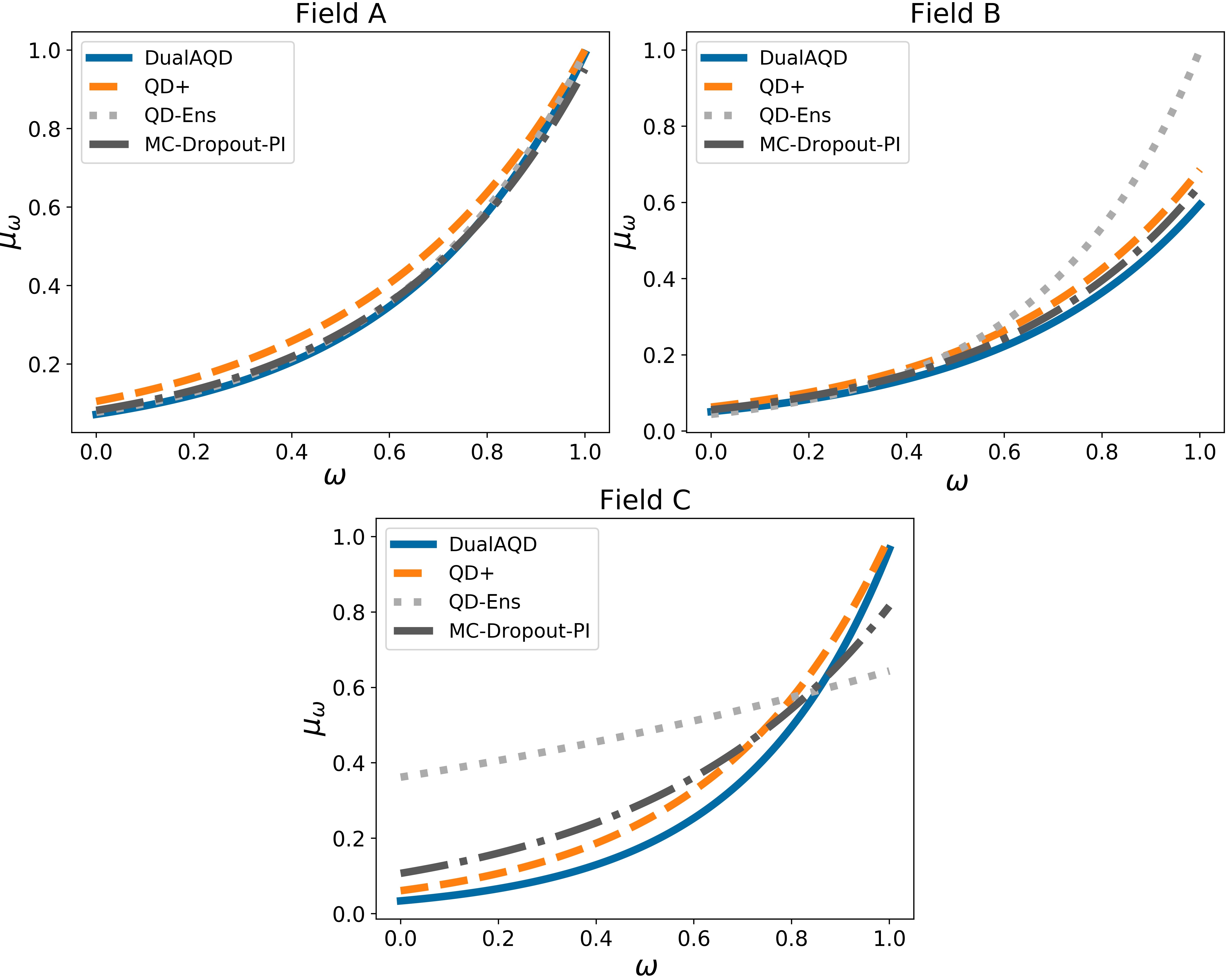}
\vspace{-4ex}
\caption{$\mu_\omega$ vs. $\omega$ comparison on yield prediction datasets. }
\label{fig:mucurves}
\end{figure}

According to the HQ principle that aims to obtain narrow PIs and high probability coverage, low $\mu_\omega$ values are preferable when comparing the performance of different PI-generation methods.
Fig.~\ref{fig:mucurves} shows the comparison of the $\mu_\omega$ metric obtained for each method on the three tested fields for different $\omega$ values.
In order to summarize the behavior shown in Fig.~\ref{fig:mucurves} into a single metric, we calculated the integral $\mu = \int_{0}^1 \mu_\omega \, d \omega$.
Since we seek to obtain low $\mu_\omega$ values for various $\omega$, low $\mu$ values are preferable.
Bold entries in Table~\ref{tab:comp1} indicate the method with the lowest $\mu$.

\begin{table}[t]
\begin{center}
\caption{PI metrics \textbf{$RMSE_{test}$}, \textbf{$MPIW_{test}$}, \textbf{$PICP_{test}$}, and $\mu$ evaluated on the yield prediction datasets.}
\label{tab:comp1}
\vspace{-2ex}
\footnotesize
\resizebox{\columnwidth}{!}{
\begin{tabular}{|c|c|c|c|c|c|}
\hline
\textbf{Field} & \textbf{Method} & \textbf{$RMSE_{test}$} & \textbf{$MPIW_{test}$} & \begin{tabular}[c]{@{}c@{}}$PICP_{test}$\\ (\%)\end{tabular} & \textbf{$\mu$} \\ \hline
\multirow{4}{*}{\textbf{A}} & 
DualAQD & 15.44 & 53.75 & 92.8 & \textbf{.350} \\ \cline{2-6} 
 & QD+ & 17.73 & 54.27 & 89.5 & .397\\ \cline{2-6} 
 & QD-Ens & 15.55 & 53.99 & 92.3 & .359\\ \cline{2-6} 
 & MC-Dropout-PI & 15.27 & 51.68 & 91.8 & .355\\ \Xhline{3\arrayrulewidth}  
\multirow{4}{*}{\textbf{B}} & 
DualAQD & 11.16 & 43.45 & 94.9 & \textbf{.221}\\ \cline{2-6} 
 & QD+ & 11.83 & 50.17 & 93.7 & .261\\ \cline{2-6} 
 & QD-Ens & 12.95 & 73.09 & 95.6 & .306\\ \cline{2-6} 
 & MC-Dropout-PI & 10.83 & 47.18 & 94.4 & .241\\ \Xhline{3\arrayrulewidth}
\multirow{4}{*}{\textbf{C}} & 
DualAQD & 18.48 & 59.96 & 96.6 & \textbf{.279}\\ \cline{2-6} 
 & QD+ & 22.27 & 62.02 & 93.9 & .336\\ \cline{2-6} 
 & QD-Ens & 17.75 & 39.93 & 63.8 & .490\\ \cline{2-6} 
 & MC-Dropout-PI & 17.15 & 50.61 & 89.3 & .349\\ \hline
\end{tabular}
}
\end{center}
\end{table}

\section{Discussion} \label{discussion}

Our loss function $Loss_{DualAQD}$ was designed to minimize the estimation error and produce narrow PIs simultaneously while using constraints that maximize the coverage probability inherently.
From Tables~\ref{tab:synth} and \ref{tab:bench}, we note that DualAQD consistently produced significantly narrower PIs than the compared methods, according to the paired $t$-test performed at the 0.05 significance level, except for the Protein dataset, where QD+ obtained comparable PI widths. 
Simultaneously, we yielded $PICP_{val}$ values of at least 95\% and better or comparable $MSE_{val}$ values. 
In addition, the ${PI_\delta}_{val}$ values reported in Table~\ref{tab:synth} demonstrate that DualAQD is the method that best adapted to the highly varying uncertainty levels of our synthetic dataset.
Thus, the PI bounds generated by DualAQD were the closest to the ideal 95\% PIs.

Notice that DualAQD obtains lower $MSE_{val}$ values than QD+ consistently despite the fact that QD+ also includes an objective function that minimizes the error of the target predictions.
The reason is that our method uses a NN (i.e., $f(\cdot)$) that is specialized in generating accurate target predictions, and its optimization objective does not compete with others.
Conversely, QD+ uses a loss function that balances four objective functions: minimizing the PI widths, maximizing PI coverage probability, minimizing the target prediction errors, and ensuring PI integrity.
The NN used by QD-Ens, on the other hand, only generates the upper and lower bounds of the PIs.
The target estimate is then calculated as the central point between the PI bounds.
As a consequence of not using a NN specialized in minimizing the target prediction error, QD-Ens achieved the worst $MSE_{val}$ values of the compared methods, except for the Year dataset.

It is worth mentioning that one of the advantages of using DualAQD over QD+ and QD-Ens is that we achieved better PIs while requiring less computational complexity.
That is, our method requires training only two NNs and uses MC-Dropout to account for the model uncertainty while QD+ and QD-Ens require training ensembles of five NNs.
In addition, QD+ requires extra complexity given that it uses a split normal aggregation method that involves an additional fitting process for each data point during testing.
Note that using deep ensembles of $M$ models is expected to perform better or similar to MC-Dropout when using $M$ forward passes~\cite{deepensembles}.
In other words, using an ensemble of five NNs, as QD and QD+ do, is expected to perform better than using five forward passes through the NN using MC-Dropout.
Nevertheless, during inference, we are able to perform not only five but 100 passes through the NN without significantly adding computationally cost.
Our method becomes more practical in the sense that, even when it uses the rough estimates of model uncertainty provided by MC-Dropout, it is still able to generate significantly higher-quality PIs.

In Fig.~\ref{fig:Power_curves}, we see the effect of using different scale factors $\eta$ to update the balancing coefficient $\lambda$ of $Loss_{DualAQD}$.
Notice that DualAQD produced wide PIs at the beginning of the training process in order to ensure PI integrity; as a consequence, the $PICP_{train}$ and $PICP_{val}$ values improved drastically.
Once the generated PIs were wide enough to cover most of the samples in the training set (i.e., $PICP_{train} \approx 1$), DualAQD focused on reducing the PI widths until $PICP_{train}$ reached the nominal probability coverage $\alpha$.
The rate at which $PICP$ and $MPIW$ were reduced was determined by the scale factor $\eta$.

Furthermore, Fig.~\ref{fig:Power_curves}a ($\eta = 0.01$) and Fig.~\ref{fig:Power_curves}b ($\eta = 0.1$) show that 
both models converged to a similar $MPIW_{val}$ value ($\sim 15$) despite having improved at different rates.
It is worth noting that we did not find a statistical difference between the results produced by the different $\eta$ values that were tested on all the datasets (i.e., $\eta \in [0.001, 0.1]$), except for the case of Kin8nm.
When various $\eta$ values were considered equally as good for a given dataset, we selected the $\eta$ value that yielded the lowest average $MPIW_{val}$, which was $\eta=0.01$ for Boston, Concrete, and Yacht, $\eta=0.005$ for Kin8nm, and $\eta=0.05$ for the rest of the datasets.
This is significant because it shows that the sensitivity of our method to the scale factor $\eta$ is low, unlike the hyperparameters required by QD-Ens, as explained in detail in Sec.~\ref{sec:bench_exp}.
What is more, our method requires a single hyperparameter, $\eta$, while QD-Ens requires two: $\lambda$ and a softening factor used to enforce differentiability of its loss function; and QD+ requires four: $\lambda_1$, $\lambda_2$, and $\lambda_3$, and the same softening factor used by QD-Ens.
Note that our method does not need an additional softening factor given that the functions of DualAQD are already differentiable.

We see in Table~\ref{tab:comp1} that DualAQD yielded better $PICP_{test}$ values than the other methods, except for field ``B'' where QD-Ens had the highest $PICP_{test}$ value, albeit at the expense of generating excessively wide PIs.
What is more, Fig.~\ref{fig:mucurves} shows that, in general, DualAQD obtained lower $\mu_\omega$ values; as a consequence, it achieved the lowest $\mu$ value in each of the three fields (Table~\ref{tab:comp1}), which implies that it offers a better width-coverage trade-off in comparison to the other methods.  
Notice that Table~\ref{tab:comp1} 
shows $PICP_{test}$ values lower than 95\% for field A.
During training and validation, the coverage probability did reach the nominal value of 95\%.
Note that, since the distribution of the test set (2020) differs from the one seen during training (2016 and 2018), the $PICP_{test}$ values may not be equal to those obtained during training. 
This illustrates the ability to show increased uncertainty when insufficient data is available for making reliable predictions.

Fig.~\ref{fig:yield} shows that DualAQD was able to produce better distributed PIs for field ``A'' (i.e., with a wider range of values) while achieving slightly better $PICP_{test}$ and $MPIW_{test}$ values than QD-Ens.
This means that DualAQD is more dynamic in the sense that it outputs narrower PIs when it considers there is more certainty and wider PIs when there is more uncertainty (recall the behavior in Fig.~\ref{fig:synth}).
As a consequence, 54.4\%, 44.3\%, and 40.3\% of the points processed by DualAQD on field ``A'' have smaller PI width than QD+, QD, and MC-Dropout, respectively, while still being able to cover the observed target values.
Similarly, 88.7\%, 65.3\%, and 49.9\% of the points processed by DualAQD on field ``B'' have smaller PI width than QD+, QD, and MC-Dropout while still covering the observed target values;
and 62.5\%, 6.0\%, and 8.8\% of the points processed by DualAQD on field ``C'' have smaller PI width than QD+, QD, and MC-Dropout while still covering the observed target values.

Finally, Fig.~\ref{fig:yield} shows that DualAQD indicates higher uncertainty in the lower (southern) region of the field, which received a nitrogen rate value that was not used in previous years (i.e., it was not available for training).
Similarly, regions of high yield values are related to high nitrogen rate values; however, there exist considerably fewer training samples of this type, which logically would lead to greater uncertainty.
Thus, there is more uncertainty when predicting regions that received high nitrogen rate values, and this is represented effectively by the uncertainty map generated by DualAQD but not the compared methods.
It is worth mentioning that even though DualAQD showed some degree of robustness empirically when given previously unseen samples, neural network-based PI generation methods do not offer any guarantee for the behavior of the model for out-of-distribution samples.  

\section{Conclusion} \label{conclusion}

Accurate uncertainty quantification is important to increase the reliability of deep learning models in real-world applications that require uncertainty to be addressed.
In this work, we focus on methods that generate prediction intervals using conventional deep neural networks for regression tasks.
As such, we presented a method that uses two companion NNs: one that specializes in generating accurate target estimations and another that has two outputs and is trained using a novel loss function designed to generate accurate and narrow PIs.

We tested our method, DualAQD, with a challenging synthetic dataset and seven benchmark datasets using feedforward neural networks. 
We also experimented with a real-world application of 2D regression using spatially correlated data to convey the usefulness and applicability of our PI generation method.
Therefore, we conclude that by using our loss function $Loss_{DualAQD}$, we were able to produce higher-quality PIs in comparison to QD+, QD-Ens, and MC-Dropout-PI; that is, our method generated significantly narrower PIs while maintaining a nominal probability coverage without detriment to its target estimation accuracy.
DualAQD was also shown to be more dynamic in the sense that it better reflects varying levels of uncertainty within the data.
It is important to point out that we achieved better performance metrics than the competing algorithms using less computational complexity and fewer tunable hyperparameters.
In the future, we plan to adapt our loss function for its use in Bayesian neural networks. 



\section*{Acknowledgments}
The authors wish to thank the members of the Data Intensive Farm Management project (USDA-NIFA-AFRI 2016-68004-24769 and USDA-NRCS NR213A7500013G021) for their comments through the development of this work, especially Dr. Paul Hegedus for collecting and curating the site-specific data. 
We thank Jordan Schupbach for providing advice on the experimental design.
 \bibliographystyle{IEEEtran}
\bibliography{main}



\begin{IEEEbiography}[{\includegraphics[width=1in,height=1.15in,clip,keepaspectratio]{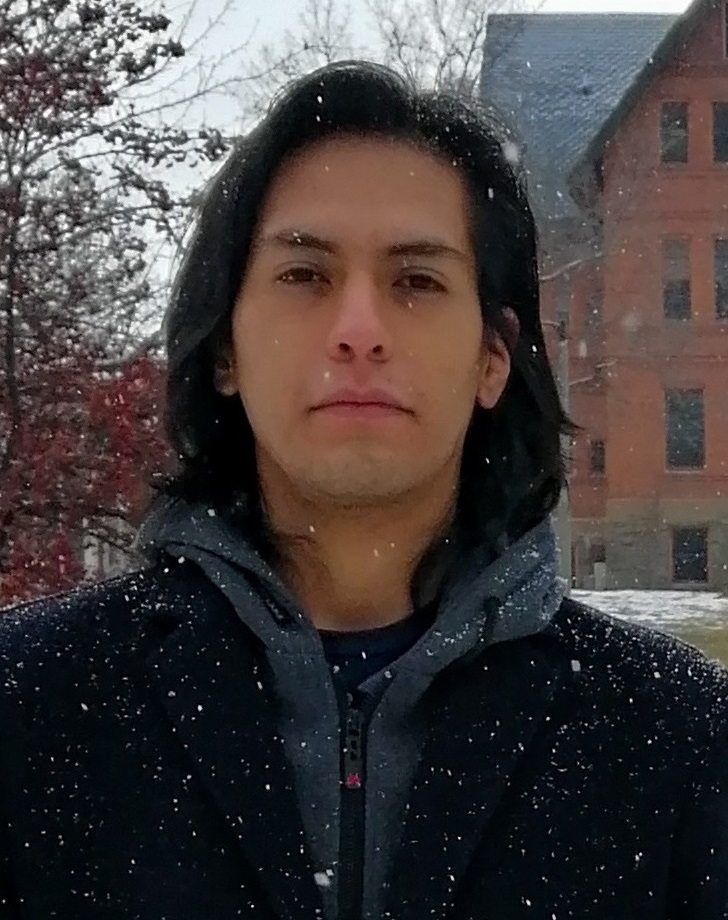}}]{Giorgio Morales}
(GS'21) is a PhD student in computer science at Montana State University and a current member of the Numerical Intelligent Systems Laboratory (NISL). 
He holds a BS in mechatronic engineering from the National University of Engineering, Peru, and an MS in computer science from Montana State University, USA. 
His research interests are deep learning, explainable machine learning, computer vision, and precision agriculture.
\end{IEEEbiography}


\begin{IEEEbiography}[{\includegraphics[width=1in,height=1.15in,clip,keepaspectratio]{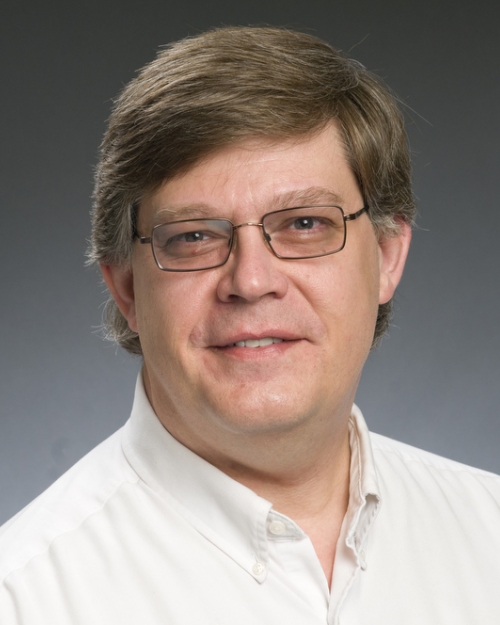}}]{John W. Sheppard} 
(M’86–SM’97–F’07) received his PhD in computer science from Johns Hopkins University and is
a fellow of the Institute of Electrical and Electronics Engineers. He is a Norm Asbjornson
College of Engineering distinguished professor of computer science in the Gianforte School
of Computing of Montana State University. His research interests include extending and applying algorithms in deep learning, probabilistic graphical models, and evolutionary optimization to
a variety of application areas, including electronic prognostics and health management, precision
agriculture, and medical diagnostics.
\end{IEEEbiography}

\vfill

\end{document}